\documentclass[letterpaper, 10 pt, conference]{ieeeconf}  

\IEEEoverridecommandlockouts    

\usepackage{graphics} 
\usepackage{amsmath} 
\usepackage{amssymb}  
\usepackage{graphicx} 
\usepackage{multirow} 
\usepackage[labelformat=simple]{subcaption}
\usepackage{xcolor}
\usepackage{mathtools}
\usepackage[ampersand]{easylist}
\usepackage{boldline}
\usepackage{booktabs}
\usepackage{adjustbox}
\usepackage{mathtools}
\usepackage{algorithm}
\usepackage[noend]{algpseudocode}
\usepackage{hyperref}
\usepackage{xspace}

\newcommand{\reffig}[1]{Fig.~\ref{#1}}

\newcommand{\reftab}[1]{Table~\ref{#1}}
\newcommand{\refsec}[1]{Section~\ref{#1}}
\newcommand{\etal}{\textit{et al.}\xspace}

\newcommand\tableheader[2]{%
  \multicolumn{1}{c}{\parbox{#1}{\centering #2}}
}

\DeclareMathOperator*{\argmin}{\arg\!\min} 

\renewcommand{\deg}{^{\circ}}
\newcommand{\norm}[1]{\left\lVert #1 \right\rVert}

\renewcommand{\vec}[1]{\mathbf{#1}}
\newcommand{\mat}[1]{\mathbf{#1}}

\algnewcommand\algorithmicforeach{\textbf{for each}}
\algdef{S}[FOR]{ForEach}[1]{\algorithmicforeach\ #1\ \algorithmicdo}

\renewcommand\algorithmicdo{}
\makeatletter
\renewcommand{\ALG@beginalgorithmic}{\footnotesize}
\makeatother

\newlength{\tempdima}
\newcommand{\rowname}[1]
{\rotatebox{90}{\makebox[\tempdima][c]{#1}}}

\title{\LARGE \bf
    Safe Local Exploration for Replanning in \\Cluttered Unknown Environments for Micro-Aerial Vehicles
}

\author{Helen Oleynikova, Zachary Taylor, Roland Siegwart, and Juan Nieto\\
Autonomous Systems Lab, ETH Z{\"u}rich%
\thanks{The research leading to these results has received funding from armasuisse, National Center of Competence in Research (NCCR) Robotics, Swiss State Secretariat for Education, Research and Innovation (SERI) n.15.0029, and the European Community’s Seventh Framework Programme (FP7) under grant-agreement n.608849 (EuRoC), n.644227 (Flourish).}
}
\begin{document}

\maketitle
\thispagestyle{empty}
\pagestyle{empty}

\begin{abstract}
In order to enable Micro-Aerial Vehicles (MAVs) to assist in complex, unknown, unstructured environments, they must be able to navigate with guaranteed safety, even when faced with a cluttered environment they have no prior knowledge of.
While trajectory optimization-based local planners have been shown to perform well in these cases, prior work either does not address how to deal with local minima in the optimization problem, or solves it by using an optimistic global planner.

We present a conservative trajectory optimization-based local planner, coupled with a local exploration strategy that selects intermediate goals.
We perform extensive simulations to show that this system performs better than the standard approach of using an optimistic global planner, and also outperforms doing a single exploration step when the local planner is stuck.
The method is validated through experiments in a variety of highly cluttered environments including a dense forest. These experiments show the complete system running in real time fully onboard an MAV, mapping and replanning at 4 Hz.
\end{abstract}

\section{Introduction}

Micro-Aerial Vehicles (MAVs) have the potential to perform many mapping and inspection missions for search and rescue and other humanitarian operations, where it is dangerous or impractical for humans to go.
Planning is a key part of any autonomous system, and online local replanning allows for fast reactions to newly observed or dynamic parts of the environment.
And while local replanning has also been recently addressed in literature, most work is shown on very low-density scenes, and makes optimistic assumptions about the environment (for example, that unknown space can be treated as free before observing it) \cite{chen2016online, pivtoraiko2013incremental}.

\begin{figure}[tb]
  \centering
  \includegraphics[width=1.0\columnwidth,trim=0 0 0 0 mm, clip=true]{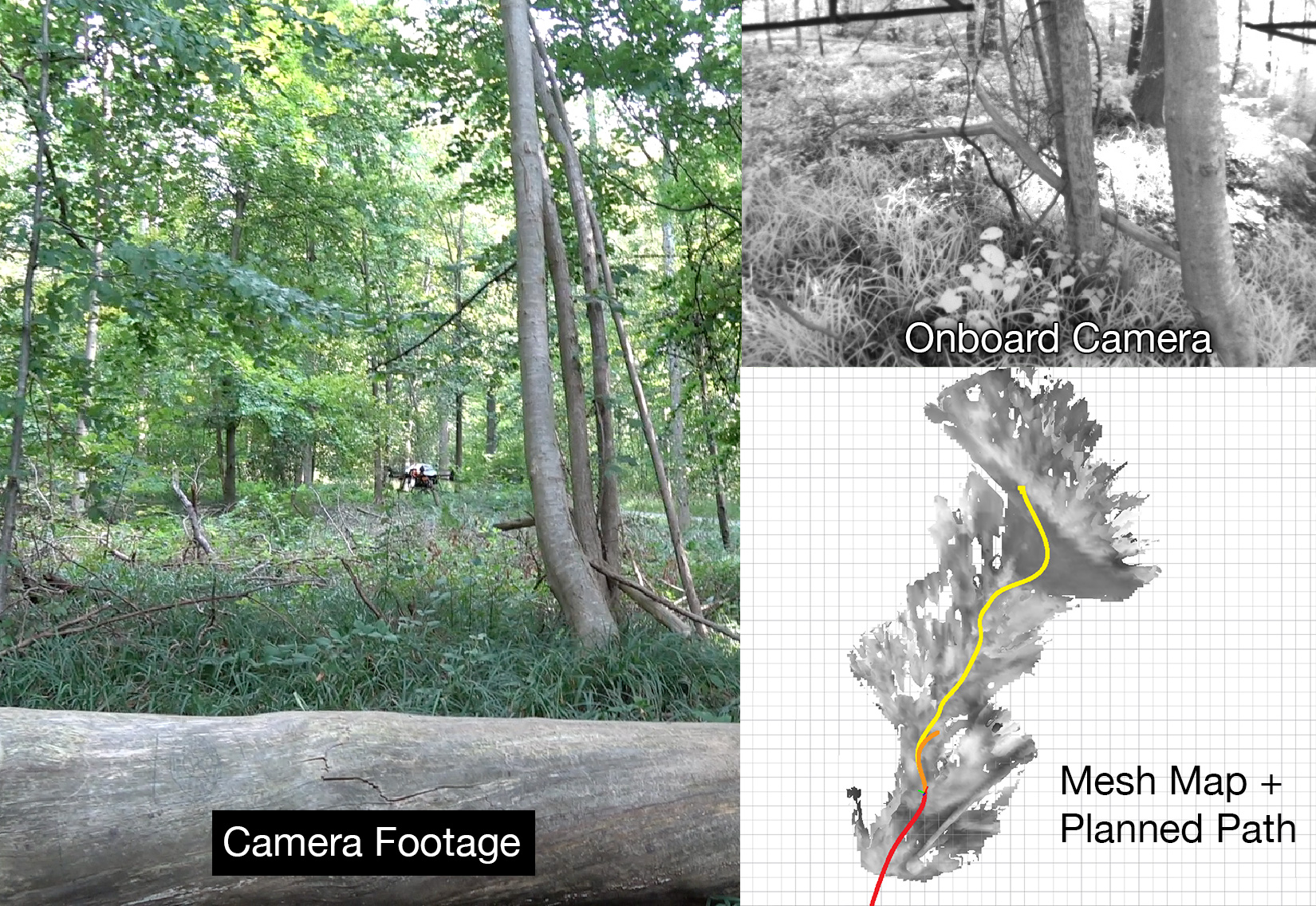}
  \caption{Experimental results from a flight through a dense forest, with video camera footage on the left, on-board view from one of the stereo cameras on the upper right, and a representation of the final mesh map on the bottom right. The final flown path is shown in yellow, the current pose of the MAV in the photos is shown as colored axes, and the planned path at the time of the photo is shown in orange.}
  \vskip-2ex
  \label{fig:forest_exp}
\end{figure}

However, in more cluttered, unknown environments, these assumptions may lead to poor planning results. Executing these plans can also be dangerous, both for the MAV and nearby people.
For example, assuming unknown space is free in forest spaces can lead to planning directly upwards into the tree canopy, which can occur as obstacles directly above an MAV are often outside the field of view of its sensors.
Alternatively if a highly conservative local planner is employed, many cluttered environment will result in the system finding no feasible paths to the goal.
In this work, we present a system that combines a conservative local planner with a local exploration strategy to navigate a cluttered, unknown environment such as the forest in \reffig{fig:forest_exp}.


Different local optimization methods for avoidance have been recently covered in literature \cite{oleynikova2016continuous-time, usenko2017real, dong2016motion}.
However, most do not explicitly address the problem of getting stuck in local minima. 
This poses a special problem in unexplored or partially unexplored environments, where only locally-optimal or reactive planners will frequently fail to find a path.
Other approaches use an optimistic global planner (one that considers unknown space as free) to overcome the problem of occasionally getting stuck.
While this works well in low-density environments, our work aims to show that this strategy (using an optmistic RRT*~\cite{karaman2011sampling} for goal selection) is not effective for highly-cluttered, partially unexplored environments.

Instead, we bring in concepts from the exploration literature to the area of local replanning.
We compare optimistic global planning to performing an exploration step from the exploration-gain-based ``next-best-view" planner (NBVP) when the trajectory optimization planner fails to find a feasible solution \cite{bircher2016receding}.
We then propose our own local exploration method, which tightly couples the local planning algorithm with a strategy that selects an intermediate goal.
The method maximizes both coming closer to the final goal and potential exploration gain, increasing the chances of finding a feasible path.

To solve the problem of map representations, our method also uses an incrementally-built, dynamically-growing Euclidean Signed Distance Field (ESDF) to compute collision costs and gradients.
The ESDF is built from a Truncated Signed Distance Field (TSDF)~\cite{oleynikova2017voxblox}, and allows us to plan in initially unknown environments with no prior knowledge of upper bounds on map size, and does not require pre-computing the object distances in batch.

We compare different parameters for our underlying local optimization method, which is an extension of our previous work~\cite{oleynikova2016continuous-time}, when the map is initially unknown, and then compare the success rates of various intermediate goal-finding strategies in highly cluttered environments.
We then demonstrate our complete system running in real-time on-board an Asctec Firefly MAV and navigating without any prior map knowledge through both an office environment and a dense forest.

The contributions of this work are as follows:
\begin{easylist}[itemize]
  & Extension of optimization problem for continuous-time polynomial trajectory optimization.
  & A system, including mapping and planning, which conservatively handles unknown space and is able to grow the map over time.
  & An active \textit{local} exploration strategy for overcoming local minima even in unknown environments by finding intermediate goal points.
  & Simulation benchmarks and real-world experiments in various cluttered environments.
\end{easylist}

\section{Related Work}
\label{sec:related}
While a large number of methods exist for local avoidance, we will address methods in 3 categories.
The first is purely reactive methods, which do not build a map of the environment but instead plan directly in the current sensor data.
While these methods are very fast and computationally efficient, they do not work well in cluttered environments where avoidance maneuvers may be non-trivial, and suffer heavily from falling into local minima.
The second class is map-based local avoidance methods, which use various techniques to compute feasible and locally-optimal paths through local maps built from sensor data or \textit{a priori} known global maps.
The last class of work we will examine here does not focus on obstacle avoidance, but instead on maximizing exploration coverage of unknown environments.
While planning collision-free paths is also a requirement for any exploration strategy, the focus is on minimizing unknown space in the final map.
We will draw inspiration from some of these methods to overcome the shortcomings of using optimization-based local planners alone.

\subsection{Reactive Avoidance}
Reactive methods focus on reacting to incoming sensor data as quickly as possible, and so act directly on obstacles in the current sensor field of view without building persistent maps.

For instance, our previous reactive work shows a method to directly convert incoming disparity maps from stereo into object segmentations, and then uses wall-following algorithm to avoid them~\cite{oleynikova2015reactive}.
Florence \etal directly integrates the nearest obstacle from a disparity map into a controller that is an open-loop library of motion primitives~\cite{florence2016integrated}. Only inexact, local state estimation is required for this approach, and they demonstrate it in both extensive simulation and real-world experiments.
Lopez and How build a kD tree of the current sensor view pointcloud, and then perform aggressive reactive avoidance from a library of fixed-velocity but variable angle motion primitives, generated from a triple-integrator model of MAV dynamics~\cite{lopez2017aggressive}.

While all three methods are shown avoiding obstacles directly in front of the MAV without prior map knowledge, they are only demonstrated on much lower obstacle densities than discussed in this paper, and suffer from not being able to avoid obstacles that are not directly in the current sensor field of view.

\subsection{Map-Based Replanning}
In contrast, most replanning methods focus on navigating in a map rather than directly on sensor data.

Richter \etal presented dynamics-aware path planning for MAVs as solving an unconstrained QP through a visibility graph generated by an RRT~\cite{richter2013polynomial}, which remains a popular method for global planning~\cite{burri2015real-time}, but is debatably too slow to replan in real-time.
Our previous work~\cite{oleynikova2016continuous-time} combines unconstrained polynomial spline optimization with gradient-based minimization of collision costs from CHOMP~\cite{ratliff2009chomp}, but is prone to local minima.
Usenko \etal utilize a similar concept, but use a B-spline representation instead, and use a circular buffer-based Octomap to overcome the issue of needing a fixed map size~\cite{usenko2017real}.
Dong \etal also use the same general problem structure as CHOMP, but represents trajectories as samples drawn from a Gaussian Process (GP) and optimize the trajectory using factor graphs and probabilistic inference~\cite{dong2016motion}.
While all these methods are able to avoid obstacles and replan in real time, none offer convincing ways to overcome the problem of getting stuck in a local minima and being unable to find a feasible solution.

Pivtoraiko \etal use graph search with motion primitives to replan online~\cite{pivtoraiko2013incremental}. However, they use an optimistic local planner: unknown space is considered traversible, and while this helps escape local minima, it is fundamentally unsafe.
Chen \etal plan online by building a sparse graph by inflating unoccupied corridors within an Octomap, then optimize an unconstrained QP to get a polynomial path~\cite{chen2016online}.
However, they only use 2D sensing and treat unknown space as free, again leading to potentially unsafe paths in very cluttered environments.

\subsection{Exploration}
The goal of exploration literature is not only to stay safe and avoid collisions, but to maximize the amount of information about the environment.
There are many different approaches, such as greedily tracking the closest unexplored frontier~\cite{heng2014autonomous} or simulating gas-like particles throughout the environment to find the sparsest area of dispersion to explore~\cite{shen2012autonomous}.

Rather than tracking frontiers, some methods instead aim to maximize information gain.
Charrow \etal optimize this gain over a state lattice with motion primitives as connecting edges, and then improve the plan with trajectory optimization~\cite{charrow2015information}.
Bircher \etal instead build an RRT tree in the unexplored space, and execute a straight-line plan to the first vertex of the most promising branch of the tree, maximizing the number of unknown voxels falling into the sensor frustum~\cite{bircher2016receding}.
Papachristos \etal extend Bircher's method by also optimizing the intermediate paths to maximize localization quality~\cite{papachristos2017uncertainty-aware}.
Similarly, Davis \etal optimize paths between next-best views to maximize coverage by introducing a coverage term to their iLQG formulation~\cite{davis2016c-opt}.

Our work combines the fast online replanning capabilities of trajectory optimization-based planning with the idea of maximizing exploration gain in a future sensor field of view.
This combination allows us to overcome the tendency of local planners to get stuck with local minima, while intelligently using our model of the system to find feasible solutions.

\section{Problem Description}
\label{sec:problem}
We aim to solve the problem of an MAV attempting to reach a goal in a previously unexplored (and completely unknown) environment.
The core focus being on very obstacle-dense and cluttered environments, with forest flight as a particular example.
The MAV has at least one 3D imaging sensor, either RGB-D or stereo, with a finite resolution and a fixed horizontal and vertical FOV, mounted in a fixed position.
We assume that the MAV is building a map of the environment from this sensor as it navigates (\refsec{sec:map}).
We design a conservative local planner, which treats unknown space as occupied and inaccessible (\refsec{sec:optimization}).
The core problem we want to address is how to design a complementary goal-finding algorithm for when the local planner gets `stuck' in a local minimum (\refsec{sec:exploration}).
All parts of the method should be fast enough to run online and in real-time entirely on-board the MAV.

\section{Local Trajectory Optimization}
\label{sec:optimization}
Our local trajectory optimization method is an extension of our previous work~\cite{oleynikova2016continuous-time}.
We represent an MAV trajectory as a high-degree polynomial spline as in Richter \etal~\cite{richter2013polynomial}, and put soft constraints (expressed in the segment time allocation) on the maximum velocity and acceleration along the trajectory, which Mellinger \etal show makes the trajectory physically feasible for a simplified dynamics model~\cite{mellinger2011minimum}.

The actual optimization minimizes a compound cost, consisting of minimizing a derivative of position such as jerk or snap as in \cite{richter2013polynomial} and \cite{mellinger2011minimum}, combined with the collision gradient cost from Ratliff \etal \cite{ratliff2009chomp}.

We will consider a polynomial trajectory in $K$ dimensions, with $S$ segments, and each segment of order $N$.
Each segment has $K$ dimensions, each of which is described by an $N$th order polynomial:
\begin{equation}
f_k(t) = a_0 + a_1 t + a_2 t^2 + a_3 t^3 \dots a_N t^N
\end{equation}
with the polynomial coefficients:
\begin{equation}
\vec{p}_k = \begin{bmatrix} a_0 & a_1 & a_2 & \dots & a_N \end{bmatrix}^\top.
\end{equation}

In order to avoid numerical issues with high orders of $t$, we instead optimize over the end-derivatives of segments within the spline \cite{richter2013polynomial}, sorted into fixed derivatives $\vec{d}_F$ (such as end-constraints) and free derivatives $\vec{d}_P$ (such as intermediate spline connections):
\begin{equation}
\vec{p} = \mat{A}^{-1} \mat{M} 
\begin{bmatrix}
\vec{d}_F \\
\vec{d}_P
\end{bmatrix}.
\end{equation}
Where $\mat{A}$ is a mapping matrix from polynomial coefficients to end-derivatives, and $\mat{M}$ is a reordering matrix to separate $\vec{d}_F$ and $\vec{d}_P$.

The final form of the optimization problem is:
\begin{equation}
\vec{d}_P^* = \argmin_{\vec{d}_P}\; w_d J_d + w_c J_c + w_g J_g \label{eq:optimization}
\end{equation}

Where the derivative cost, $J_d$, aims to minimize a certain derivative (often jerk or snap) of the position \cite{mellinger2011minimum}, with $\mat{R}$ as the augmented cost matrix.
\begin{eqnarray}
J_d &=& \vec{d}_F^\top \mat{R}_{FF} \vec{d}_F + \vec{d}_F^\top \mat{R}_{FP} \vec{d}_P + \nonumber\\
&& \vec{d}_P ^\top \mat{R}_{PF} \vec{d}_F + \vec{d}_P^\top \mat{R}_{PP} \vec{d}_P
\end{eqnarray}


The collision cost, $J_c$, is an approximation of the line integral of costs along the path, where $c(\vec{x})$ is the collision cost from the map, $\vec{f}(t)$ is the position along the trajectory at time $t$, and $\vec{v}(t)$ is the velocity at time $t$:
\begin{equation}
J_c = \sum_{t=0}^{t_m} c (\vec{f}(t)) \norm{\vec{v}(t)} \Delta t \label{eq:j_c}
\end{equation}
We use 3 segments and optimize jerk, as was found to be the best settings in our previous work~\cite{oleynikova2016continuous-time}.

We extend our previous work by using a soft cost for the goal, $J_g$, similarly to \cite{usenko2017real}, and the local goal finding below.
\begin{equation}
J_g = \norm{\vec{f}(t_\textrm{end}) - \vec{g}} \label{eq:j_g}
\end{equation}
where
\begin{equation}
f_k(t_\textrm{end}) = \vec{T_\textrm{end}}\mat{A}^{-1}\mat{M} \begin{bmatrix}
\vec{d}_{Fk} \\
\vec{d}_{Pk}
\end{bmatrix}
\end{equation}

%

This allows the optimization to slightly adjust the goal point to allow better trajectories, or find feasible trajectories at all.
An analysis of the effect of this term on the success rate is offered in \refsec{sec:simulation}.

In general, even with the soft cost term, the initial state of the optimization problem should have the end point be free or almost free of collisions.
In our system, we set a fixed planning horizon $r_p$, which is the maximum distance from the current state that the planner is allowed to go to.
However, projecting a global goal $\vec{g}_g$ onto the sphere of this radius often leads to occluded end points.

In \refsec{sec:simulation}, we compare two different strategies for moving this end-goal to be a feasible end point for the spline: straight-line goal finding, which backtracks along the line from the projection of $\vec{g}_g$ to the start point of the trajectory, $\vec{x}_s$, until the first unoccupied point along this line.
The second method is gradient-based in the map: from the projection of $\vec{g}_g$ onto the sphere, we evaluate the gradient of the collision cost map and follow the gradient down until a free-space location is found.
If the approach becomes stuck in a local minimum of the gradient, we evaluate the straight-line strategy for one step.

Finally, these trajectories are only planned on $\mathbb{R}^3$ and derivatives.
To map these trajectories to the full pose of the MAV on $SE^3$, pitch and roll are defined by the acceleration in $x$ and $y$ directions, while yaw $\gamma$ remains free.
We use velocity-tracking yaw to increase the chances of the MAV seeing new or dynamic obstacles before collision:
\begin{equation}
\gamma(t) = \arctan\Big(\frac{v_y(t)}{v_x(t)}\Big)
\end{equation}

\section{Map Representation and Unknown Space}
\label{sec:map}

As the optimization method in \refsec{sec:optimization} requires not only distances to the nearest obstacles but also the gradients of these distances, we require a map representation that can be efficiently queried for this information.
While our original work \cite{oleynikova2016continuous-time} used a fixed-size Euclidean Signed Distance Field (ESDF) built from an octomap representation, we more recently presented a way to build ESDFs from Truncated Signed Distance Fields (TSDFs) efficiently.
This allows the system to incrementally build maps of arbitrary size from sensor data in real time.
This system, called \textit{voxblox}\footnote{\url{github.com/ethz-asl/voxblox}}, is used as the map representation for the proposed planner~\cite{oleynikova2017voxblox}.

The map consists of both the original TSDF, built from sensor data, which contains projective signed distances to surfaces within a very small truncation distance to the object and free space information, and the ESDF which contains Euclidean distances to obstacles in a much larger radius.
The details of how to build both representations incrementally is addressed in \cite{oleynikova2017voxblox}.

To implement the desired property of treating unknown space as occupied, we modify the ESDF with data from the current state of the robot.
One critical issue with treating unknown space as occupied is that the starting position of the robot will never be observed and will always be treated as occupied.
For this reason, we change the ESDF values of unknown voxels in a small clearing radius $r_c$ around the initial pose of the MAV to free. 
$r_c$ should ideally be only slightly larger than the collision checking radius of the robot.

We also take a large radius $r_o$, which should be greater than or equal to the maximum planning radius, and set all unknown voxels in this radius to occupied.
Marking unknown as occupied is essential to conservative local planners, as allowing free entry into unknown space leads to behaviors such as slamming into the ceiling when presented with obstacles in front.

\section{Intermediate Goal Selection}
\label{sec:exploration}
In addition to the mapping and local planning methods presented above, we need an active exploration strategy to overcome the shortcomings of local trajectory optimization methods in very cluttered, partially unknown environments.
A typical solution to this problem is to use an optimistic global planner, which assumes unknown space is free, to select a new set of waypoints to track~\cite{nieuwenhuisen2016layered}.

In the results section (\refsec{sec:simulation}), we quantitatively compare five core methods of selecting new waypoint locations.
The first method is naive random waypoint selection.
When the local optimization fails, we select a new 3D waypoint position at random within a sphere of the starting position of the trajectory.
The planner then attempts to track this waypoint, until it is either reached or another infeasible solution is encountered.
Then the new waypoint is set to the original goal point.
This strategy (one random, one back to original goal) continues until either the original goal point is reached or the maximum number of replans is exceeded.

The second strategy is an optimistic (unknown = free) RRT*~\cite{karaman2011sampling} visibility graph.
This aims to best simulate the global planners used in other approaches, such as \cite{usenko2017real} and \cite{burri2015real-time} \cite{nieuwenhuisen2016layered}.
Since as \refsec{sec:map} describes, we set a large radius of unknown space to occupied in the ESDF, we instead use the raw TSDF as the obstacle map and treat unknown voxels as unoccupied.
We then generate a sparse visibility graph toward the final goal, and track the first waypoint in the graph.
If the first waypoint is reached, then we keep iterating through the graph until the goal point.
If at any time, the local planner is stuck again, we generate a new RRT* plan.

We also consider the opposite strategy: a conservative or pessimistic RRT*, which assumes unknown space is occupied.
Since the underlying local planner is also conservative, if it is unable to find a solution, it is likely that no solution to the goal exists through free space.
Therefore, we build the RRT graph and select the node in the tree that has the closest Euclidean distance to the goal point, and then track the first vertex in the branch of the tree that the closest node belongs to.

The next strategy we consider is directly from the exploration literature, the ``next-best view" planner (NBVP) from Bircher \etal~\cite{bircher2016receding}.
Their approach consists of building a rapidly-exploring random tree (RRT) with a small number of nodes in position and yaw space, and simulating the expected view fustrum of the camera sensor. The approach then selects the first node to execute in the branch that leads to the highest information gain in terms of unknown voxels that would be observed.
We implement this approach for comparison; however, since there is no goal-tracking component to this exploration strategy, we use the same scheme as with the random waypoint selection: one exploration waypoint, followed by trying to reach the goal, followed by another exploration waypoint.

The final strategy is our exploration strategy, combining aspects of both the exploration strategy above and goal-tracking and sensor field of view awareness, described in detail below.

\subsection{Proposed Method}
Our method uses a similar methodology to NBVP, where the potential exploration gain of future points is evaluated by projecting the camera frustum into the voxel grid.
However, we adapt the method to (i) better suit the purpose of increasing the chances of the robot making it to the goal, and (ii), to function online, in real-time in a high-rate loop.
The core differences are that we do not build an RRT graph, we subsample within the view frustum, we do not do raycasting to find occlusions, and we introduce a goal-seeking reward in addition to the exploration gain.

Our method works as follows: first, we draw the global goal $\vec{g}$ with some probability $P_g \in (0, 1)$.
Otherwise, we proceed to generate $N$ random points, $\vec{x}_n$, in the \textit{unoccupied} space of the TSDF, within a maximum radius $r$ of the start point of the trajectory $\vec{x}_s$.
Note, importantly, that we use the original TSDF rather than the ESDF to select these points and evaluate the frustum, as the ESDF sets nearby unknown space to occupied for planning safety purposes.

We select a yaw $\gamma$ for each point by finding the angle of the vector from the trajectory start $\vec{x}_s$ to the sampled point $\vec{x}_n$, to approximate the real velocity-facing yaw.
For each of these points, we evaluate the exploration gain of the camera frustum at that point by counting the number of unknown voxels in the TSDF.
The exploration gain function $l(\vec{x}, \gamma)$ can be expressed as:
\begin{equation}
l(\vec{x}, \gamma) = \# \{v | v \in \textrm{frustum}(\vec{x}, \gamma) \cap v \in \textrm{unknown}(v)\}
\end{equation}

\begin{figure}[tb]
  \centering
  \includegraphics[width=0.7\columnwidth,trim=0 0 0 0 mm, clip=true]{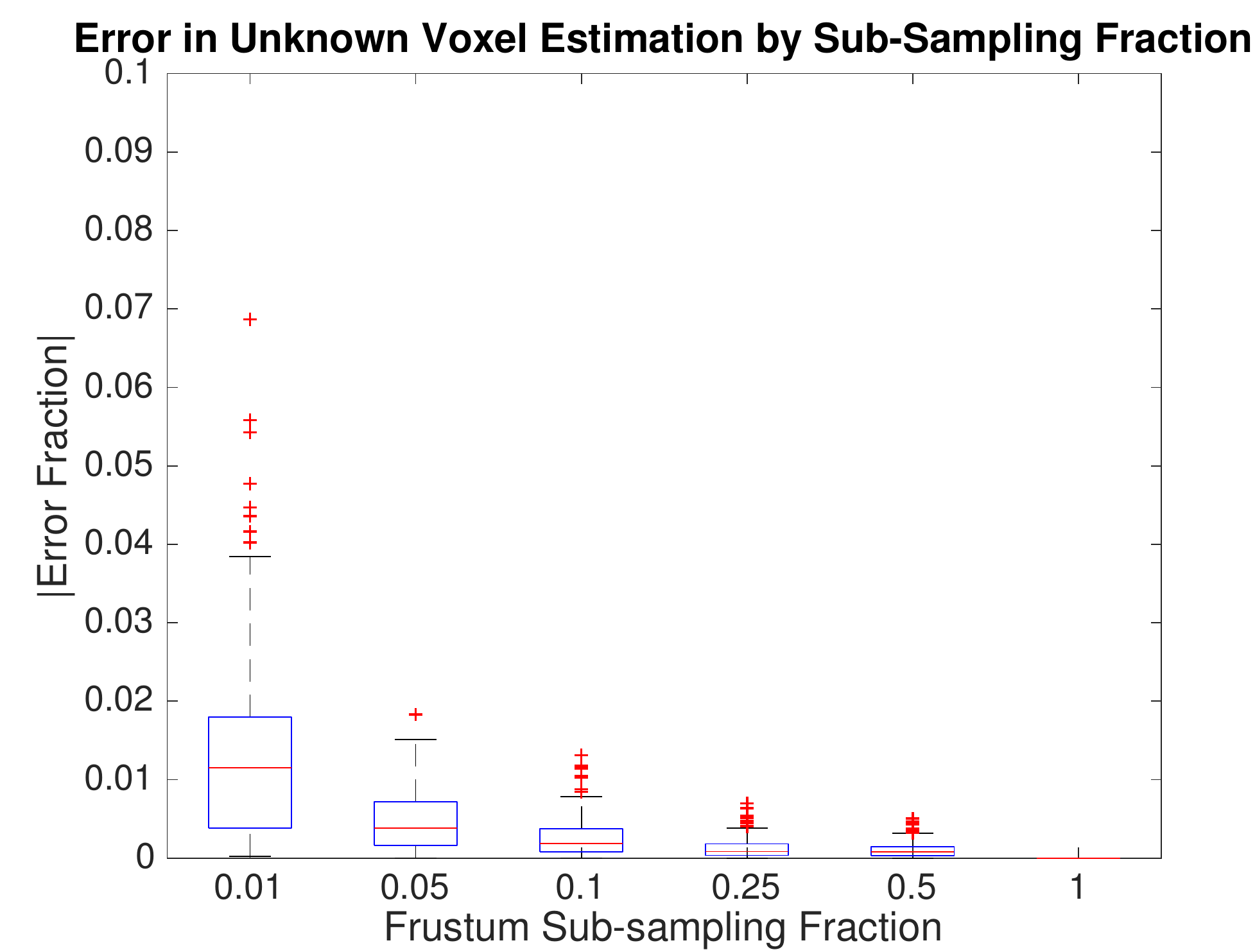}
  \caption{Error in estimation of unknown voxels in the sensor frustum (as a proxy for exploration gain), by subsampling fraction (a subsampling fraction of 0.01 = 1\% of the samples are taken). As can be seen, a sampling of 5\% of the samples yields only a maximum 2\% error in the unknown voxel estimation but could lead to up to a 20$\times$ speedup in lookup operations.}
  \label{fig:sampling}
\end{figure}

In order to run in real-time, we approximate the actual exploration gain by subsampling the frustum by a certain factor, and checking only every $s$th voxel.
We evaluate the effect of this approximation in \reffig{fig:sampling} in simulation, which shows that sampling only 5\% of the samples usually leads to an estimation error of less than 1\%, and in practice runs 3 times faster than evaluating the full frustum.

Additionally, for each point we also evaluate the distance to the global goal, normalized by the maximum distance to goal $d_g$ (to allow consistent weighting across different settings and goal distances).
This normalized distance is converted to a reward, giving the total reward function $R$ for each point $\vec{x}_n$ as:
%
%
\begin{eqnarray}
d_{g} & = & \norm{\vec{g} - \vec{x}_s} + r \\
R(\vec{x}_n, \gamma, \vec{g}) & = & w_{e} l(\vec{x}, \vec{\gamma}) + w_{g} \frac{d_g - \norm{\vec{g}-\vec{x}_n}}{d_g}
\end{eqnarray}

The point with the highest reward is chosen as the next intermediate goal.

\begin{figure}[tb]
  \centering
  \includegraphics[width=1.0\columnwidth,trim=0 0 0 0 mm, clip=true]{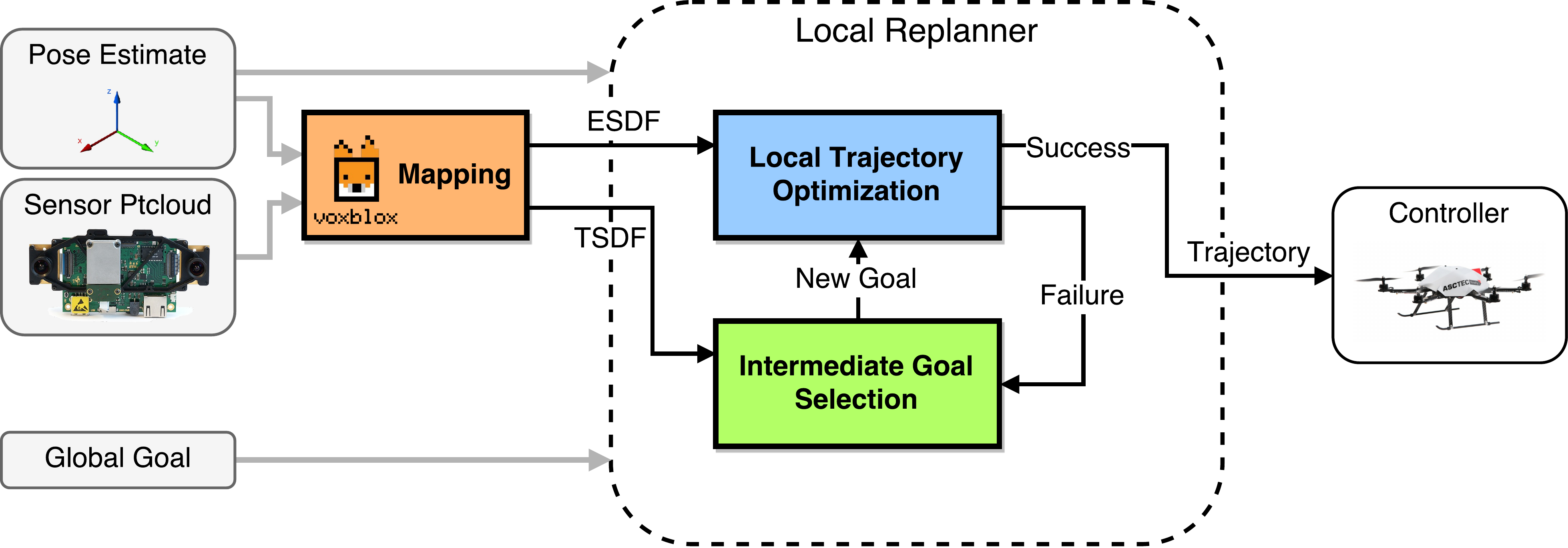}
  \caption{System diagram of the mapping and planning subsystems. The ESDF is used by the trajectory optimizer to compute collision costs, and the TSDF is used by the intermediate goal finding (local exploration algorithm) to evaluate exploration gain. If the trajectory optimization succeeds, the trajectory is sent to the controller; otherwise we attempt to find an alternative intermediate goal.}
  \label{fig:sys_diag}
\end{figure}

A diagram showing the complete system (including mapping) is shown in \reffig{fig:sys_diag}.

%

\section{Simulation Experiments}
\label{sec:simulation}
This section will evaluate different aspects of our system in a simulation environment where the ground truth map is known.
We evaluate the effect of parameters on success rate of local trajectory optimization, compare the intermediate goal finding methods presented in \refsec{sec:exploration}, and the effect of subsampling the camera view frustum for exploration gain evaluation.

These simulations are made with the \textit{voxblox}, which allows generating ground-truth ESDFs for environments made of primitive shapes (in this case, cylinders to simulate trees in a forest), and also allows simulating sensor measurements by raycasting into the map.
The maps are 15 $\times$ 10 meters, and have an obstacle region of 10 $\times$ 10, to ensure that the start and end poses are always free.
Cylinders of radii between 0.1 and 0.5 m and various heights are placed randomly within the space.
The objects per square meter metric maps to approximately to percentage of the volume occupied, $\pm 5\%$ (for instance, 0.4 objects/m$^2$ is 35-45\% occupied volume).

For the purposes of these experiments, we assume our MAV can track the polynomial trajectories perfectly, which \cite{mellinger2011minimum} shows is possible as long as we respect maximum velocity and acceleration bounds while planning.
We add a new viewpoint into the incrementally-built map and then replan once a second of simulation time.
The incremental planning methods have a maximum planning horizon of 3 meters.

\begin{figure}[tb]
  \centering
  \begin{subfigure}[b]{0.31\columnwidth}
     \centering
      \includegraphics[width=1.0\columnwidth,trim=0 0 0 0 mm, clip=true]{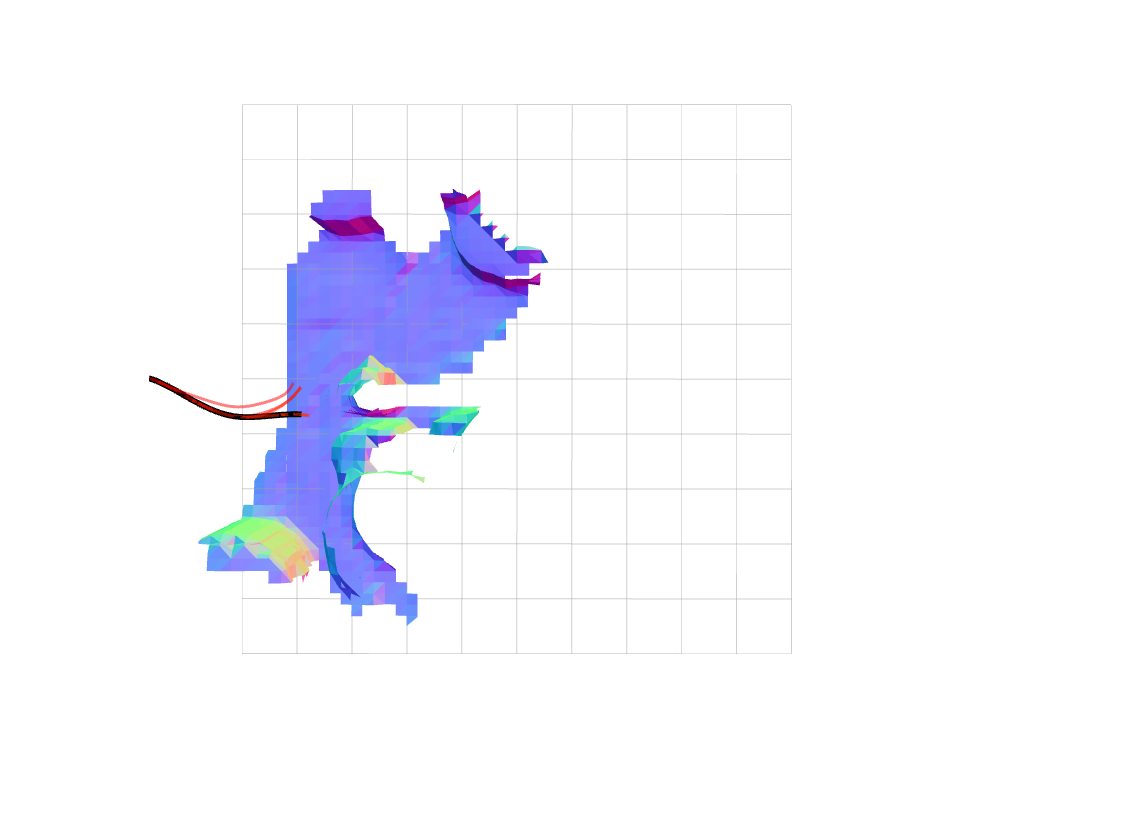}
      \caption{No Inc. Goal}
      \label{fig:no_goal}
  \end{subfigure}
  \begin{subfigure}[b]{0.31\columnwidth}
    \centering
    \includegraphics[width=1.0\columnwidth,trim=0 0 0 0 mm, clip=true]{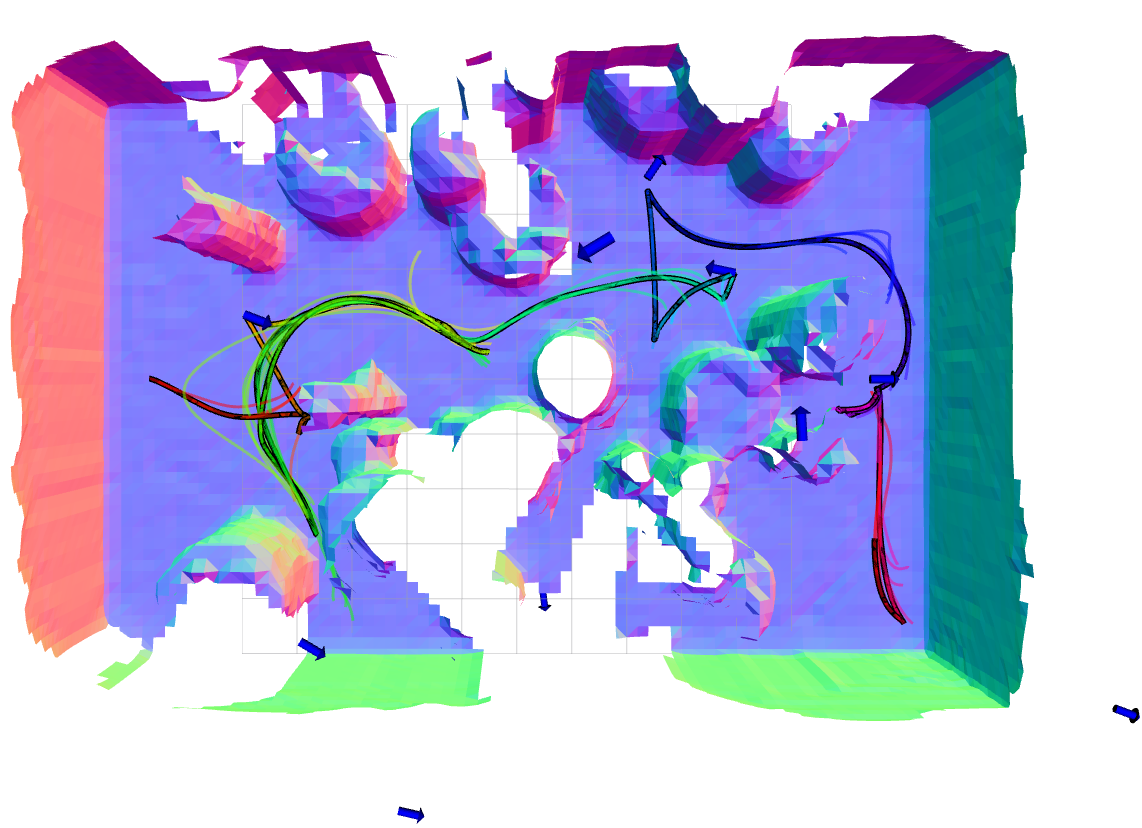}
    \caption{Random Goals}
    \label{fig:random_goals}
  \end{subfigure}
  \begin{subfigure}[b]{0.31\columnwidth}
  \centering
  \includegraphics[width=1.0\columnwidth,trim=0 0 0 0 mm, clip=true]{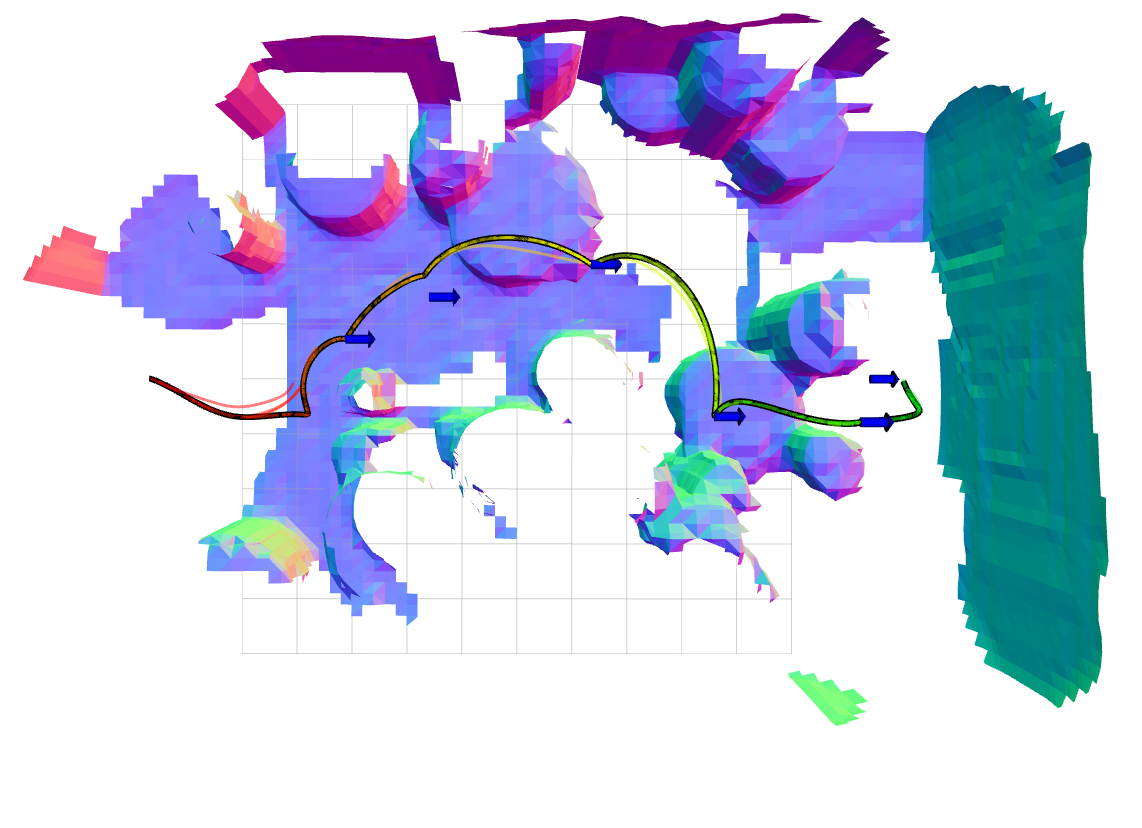}
  \caption{RRT* (Opt.)}
  \label{fig:rrt_opt}
\end{subfigure}
  \begin{subfigure}[b]{0.31\columnwidth}
  \centering
  \includegraphics[width=1.0\columnwidth,trim=0 0 0 0 mm, clip=true]{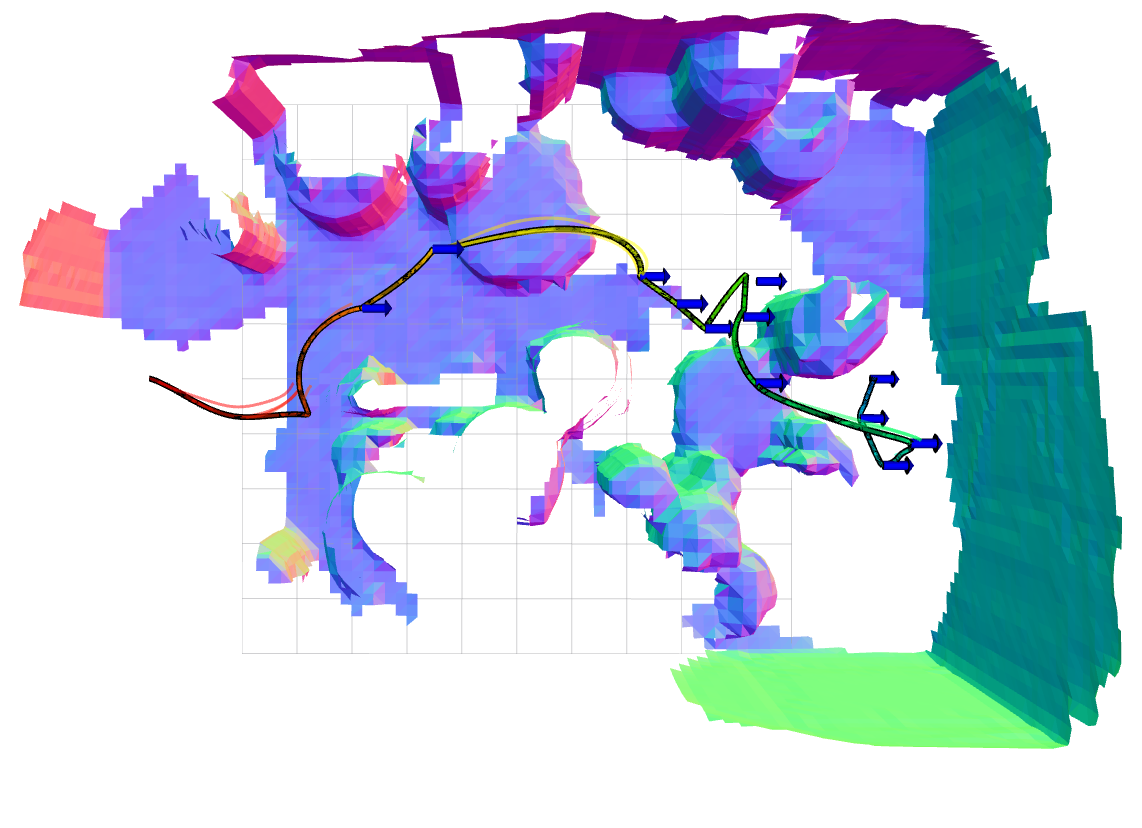}
  \caption{RRT* (Cons.)}
  \label{fig:rrt_cons}
\end{subfigure}
  \begin{subfigure}[b]{0.31\columnwidth}
  \centering
  \includegraphics[width=1.0\columnwidth,trim=0 0 0 0 mm, clip=true]{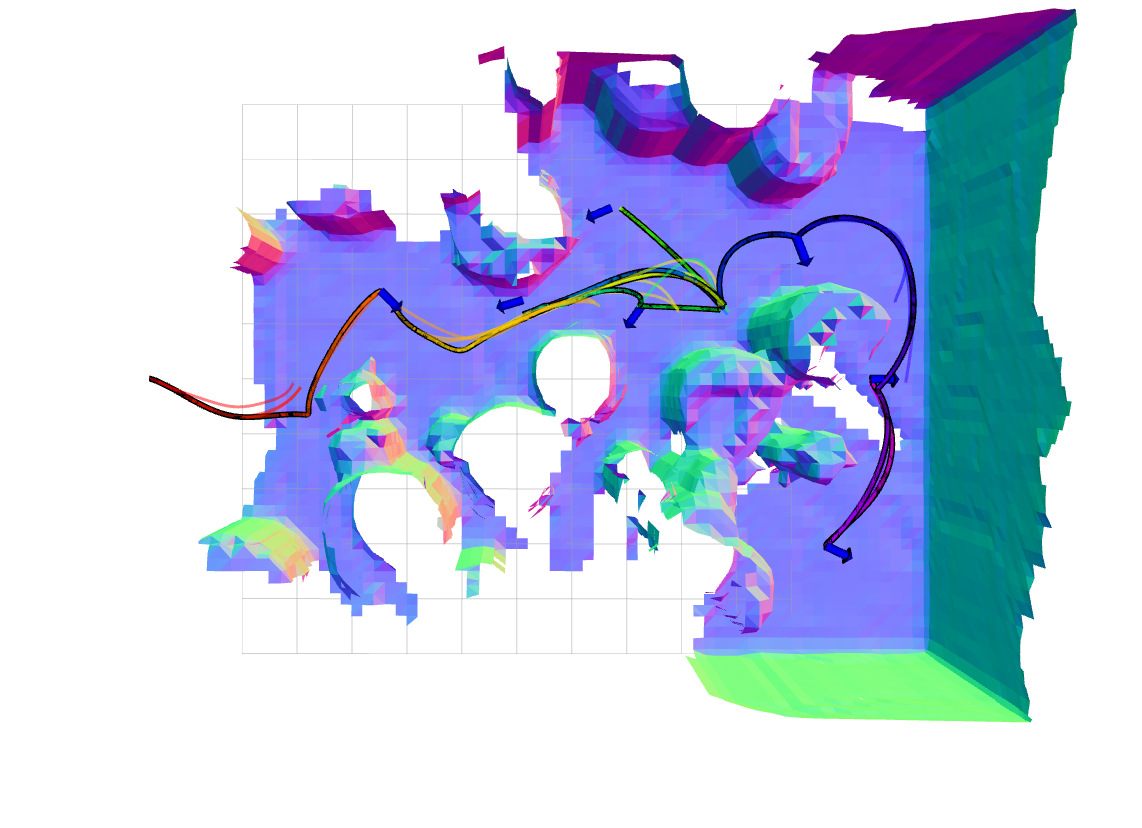}
  \caption{NBVP}
  \label{fig:nbvp}
\end{subfigure}
  \begin{subfigure}[b]{0.31\columnwidth}
  \centering
  \includegraphics[width=1.0\columnwidth,trim=0 0 0 0 mm, clip=true]{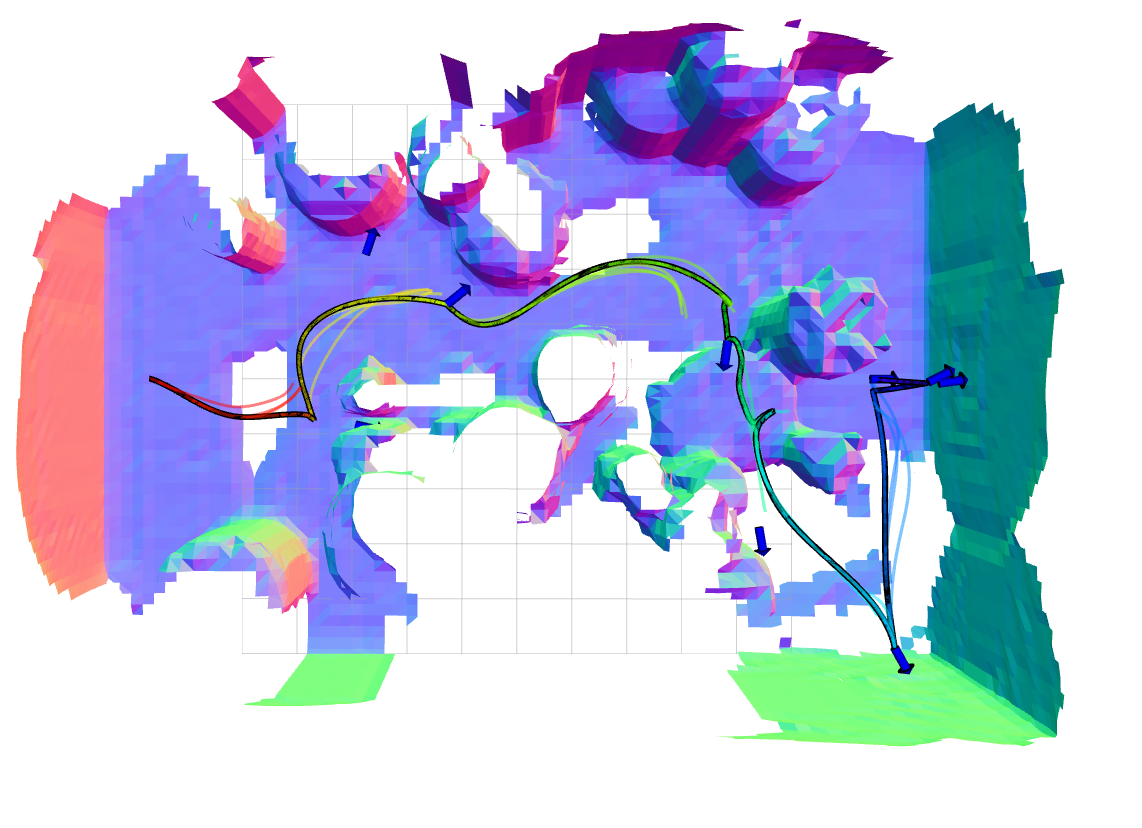}
  \caption{Our Method}
  \label{fig:ours}
\end{subfigure}
 \caption{Comparison of methods in a small simulation case (15 $\times$ 10 meters) with 0.3 objects/m$^2$ obstacle density. The black line shows the final path, the colored lines show intermediate paths, and dark blue arrows show the intermediate goals selected by the algorithm. Only our method and NBVP were successfully able to solve the case; both RRT* methods were unable to see the final location as free as they do not consider sensor field-of-view in the planning, and the random goal selection had too few replans. All methods ran for up to 120 replans.}
 \label{fig:comparison_methods}
\end{figure}

\begin{figure}[tb]
  \centering
  \includegraphics[width=0.6\columnwidth,trim=0 0 0 0 mm, clip=true]{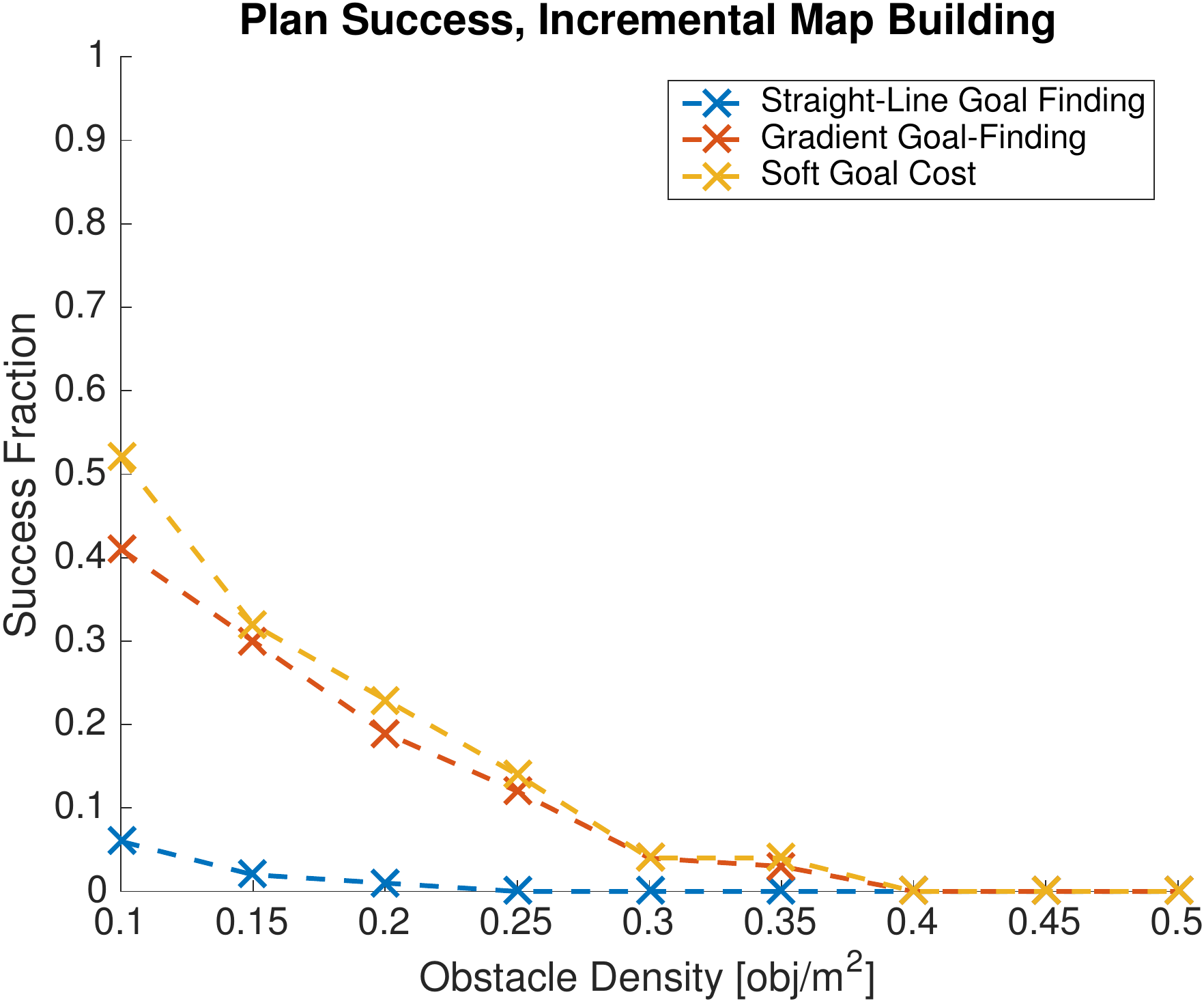}
  \caption{A comparison of the planner success without any intermediate goal-finding strategy, building the map incrementally. As can be seen, gradient-based goal-finding significantly increases success chance over line-based goal finding, and soft goal-cost further increases performance. There are 100 trials per density, with 60 replans (60 seconds at 1 Hz replanning rate).}
  \label{fig:incremental}
\end{figure}

The first results are for starting with an a completely empty map, and inserting new viewpoints along the path at 1 Hz, with a sample solutions from no incremental goal-finding shown in \reffig{fig:no_goal} and the quantitative results in \reffig{fig:incremental}.
As can be seen, straight-line goal finding and purely local optimization are able to solve only a very small percentage of the test cases.
Using gradient-based goal finding significantly increases the performance, and soft goals further increase success rate, especially at lower densities.
However, the success rates overall are still unacceptably low.

To overcome these issues, we benchmark the intermediate goal finding methods, described in \refsec{sec:exploration}, on the same simulation cases.
Example qualitative results are shown for all methods in \reffig{fig:comparison_methods}.
The simulations show the differences between the methods: random goals fails to find the goal within the allocated time as the intermediate goals are too undirected, and the two RRT*-based methods fail since they do not consider the field-of-view of the sensor and are therefore never able to observe the goal point as clear.

\begin{figure}[tb]
  \centering
  \includegraphics[width=0.6\columnwidth,trim=0 0 0 0 mm, clip=true]{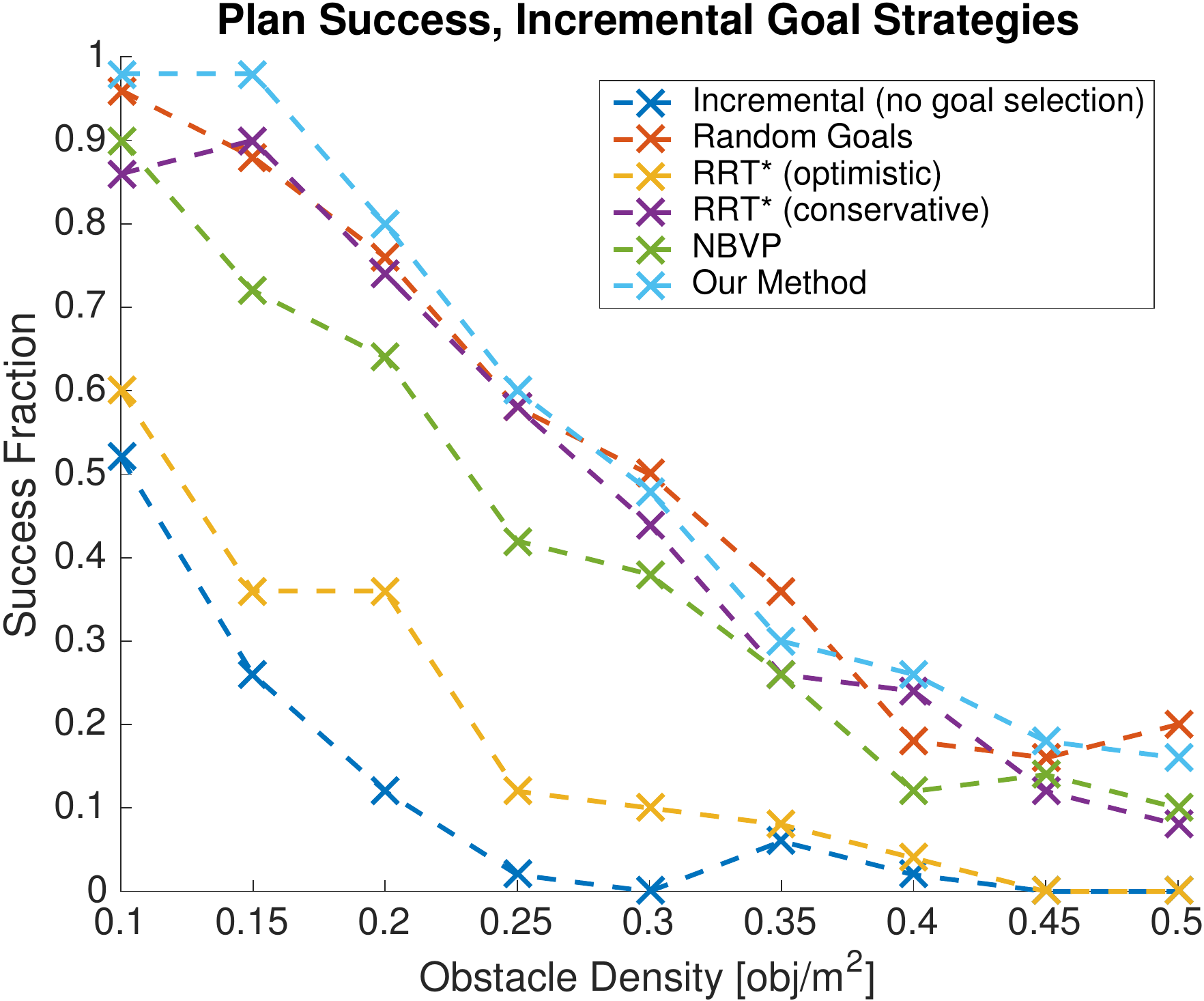}
  \caption{A comparison of the success rates for different incremental goal-finding strategies. Note that our method, NBVP, and conservative RRT* sare able to solve more test cases than optimistic RRT*, which is commonly used as a global planner. There are 50 trials per density, and a maximum of 120 replans per trial.}
  \label{fig:goal_strategy}
\end{figure}

\reffig{fig:goal_strategy} shows the quantitative results: as can be seen, all goal-finding methods outperform the naive optimization-only method.
The optimistic RRT* performs the worst, as it tends to select the same infeasible path over and over again as unknown space is marked as traversable for this method.
NBVP performs somewhat better, as it uses the sensor model to maximize exploring the small area.
Conservative RRT* performs comparatively well, as it is simply tracking the closest free point to the goal, but has no knowledge of the sensor model.

\begin{figure}[tb]
  \centering
  \includegraphics[width=0.6\columnwidth,trim=0 0 0 0 mm, clip=true]{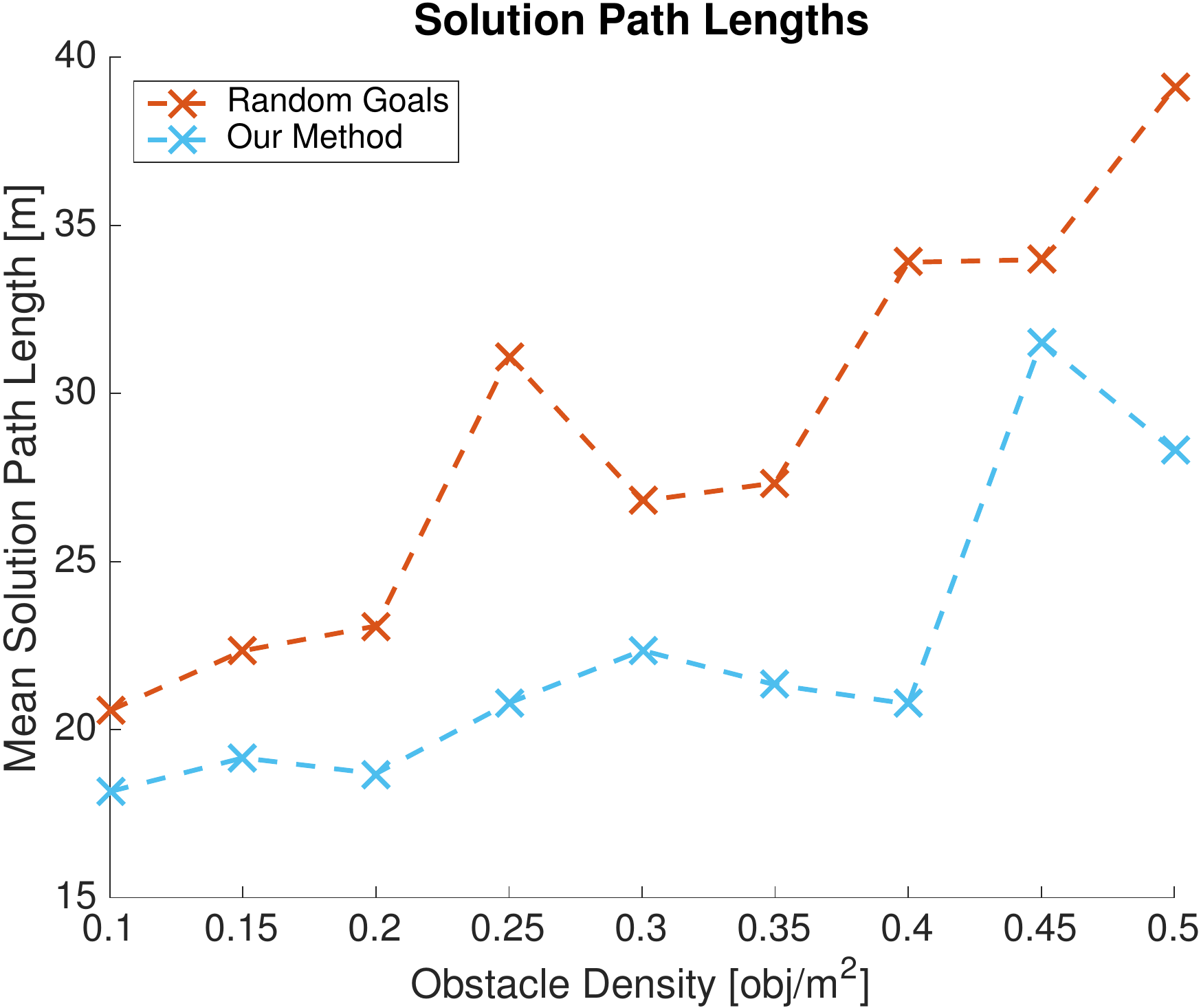}
  \caption{Path length comparison between random goal selection and our proposed method. The path lengths are only evaluated for trials where both planners succeeded, to allow a fair comparison. Note that our method always finds a solution in a significantly shorter path length, as it exploits current knowledge of the environment.}
  \label{fig:path_length}
\end{figure}

Finally, our method performs on par with random goal selection in terms of success rate.
However, our method is able to consistently produce much shorter path lengths: \reffig{fig:path_length} shows the mean path lengths for simulation cases that \textit{both} random goal finding and our method were able to solve.
Our method produces paths up to 35\% shorter.

\begin{figure}[tb]
  \centering
  \includegraphics[width=0.65\columnwidth,trim=0 0 0 0 mm, clip=true]{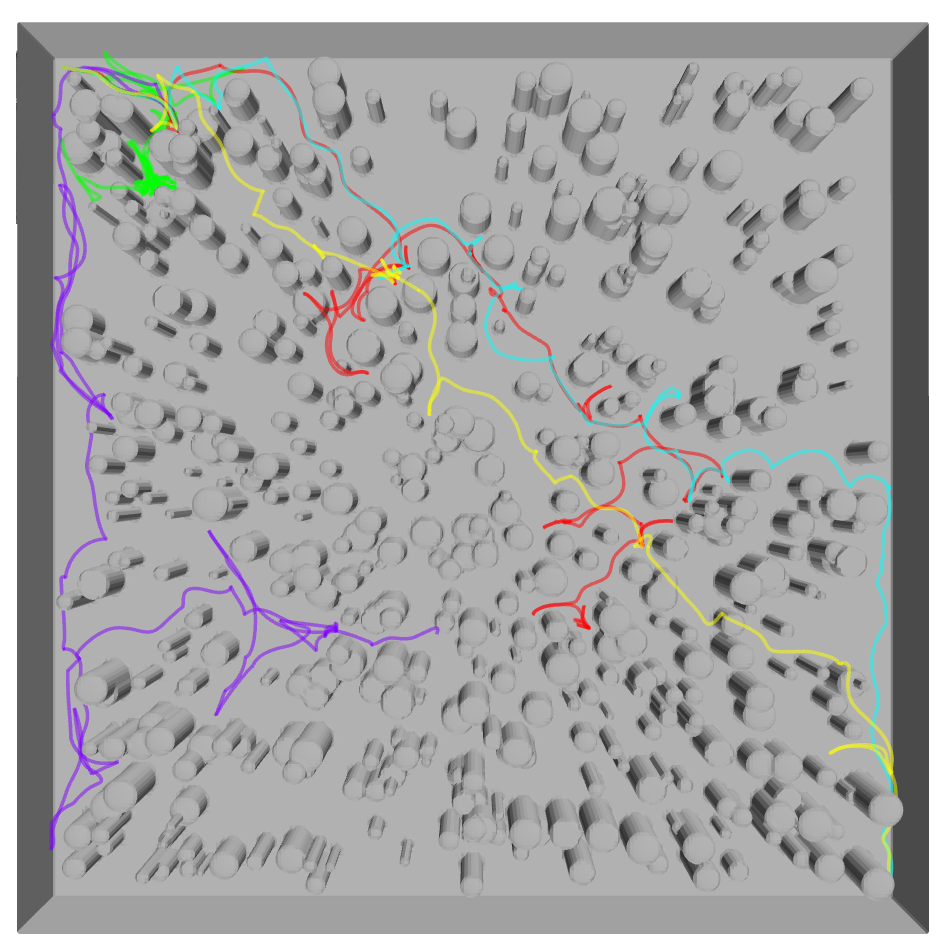}
  \caption{A 50 m $\times$ 50 m randomized ``forest" environment, with a density of 0.2 objects$/$m$^2$. The lines compare five planners: no goal selection (dark blue), random goal finding (red), optimistic RRT* (yellow), conservative RRT* (purple), NBVP (green), and our method (teal). All planners start in the upper-left corner and try to reach the lower-right, but only our planner and optimistic RRT* are able to successfully find a solution in 500 replan cycles. The planning is in 3D, so some plans go over an obstacle.}
  \label{fig:long_benchmark}
\end{figure}

\begin{figure}[tb]
  \centering
  \includegraphics[width=0.6\columnwidth,trim=0 0 0 0 mm, clip=true]{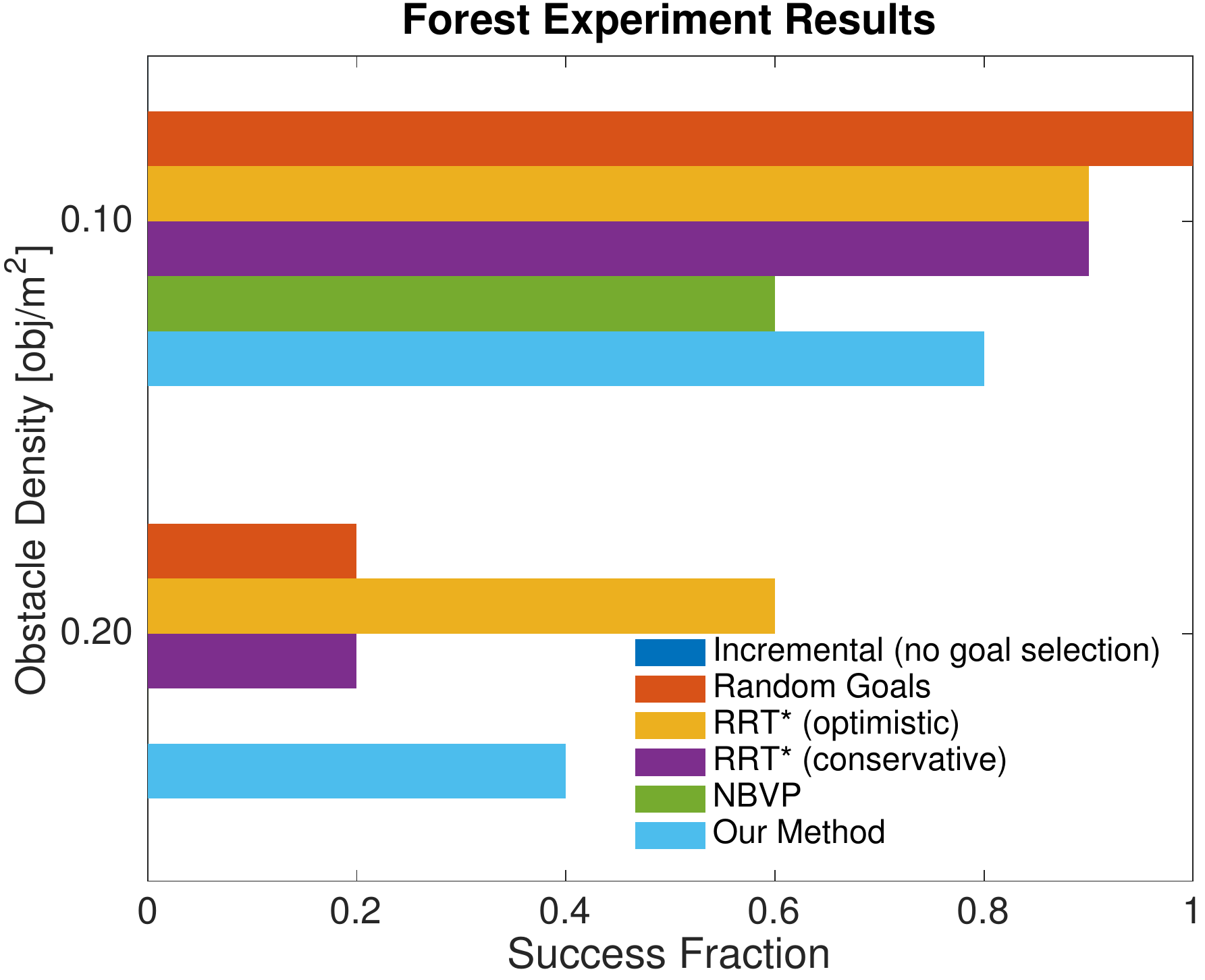}
  \caption{Quantitative results of success rate from the long forest simulation, limited to 500 replan cycles. While the RRT-based methods can offer good performance in this situation they also sample orders of magnitude more points than our method, and require much more time to select an intermediate goal.}
  \label{fig:forest_results}
\end{figure}

The final experiment is a more realistic test of a long forest traversal.
We generate a 50 meter $\times$ 50 meter randomized map with 0.1 and 0.2 obstacles/m$^2$, and set the MAV to explore from one corner to the other.
The results from 0.2 density are shown in \reffig{fig:long_benchmark}, where our method and optimistic RRT* were the only two to successfully make it to the goal.
Simulation results over 20 maps at different densities are shown in \reffig{fig:forest_results}, and the timings of different aspects of our method from this simulation are shown in \reftab{table:timing}.

%

\section{Real-World Experiments}
\label{sec:experiments}
To evaluate our system in a real-world scenario, we performed multiple experiments in two different test environments: a cluttered office space and a dense forest with a variable ground height.
The results of all described experiments are available at \url{https://youtu.be/rAJwD2kr7c0}.

All of the experiments start with a completely unknown map, use visual-inertial odometry from the forward-facing (with a 12$\deg$ downward pitch) stereo camera, update the map from stereo and replan at 4 Hz, and run everything entirely on the 2.4 GHz i7 dual-core CPU on-board the robot.
We use rovio for state estimation~\cite{bloesch2015robust}, a non-linear MPC for position control~\cite{kamel2016linear}, and the Asctec on-board attitude controller.
The average flight velocity was 1.0 m/s.

\begin{figure}[tb]
  \centering
  \includegraphics[width=1.0\columnwidth,trim=0 0 0 0 mm, clip=true]{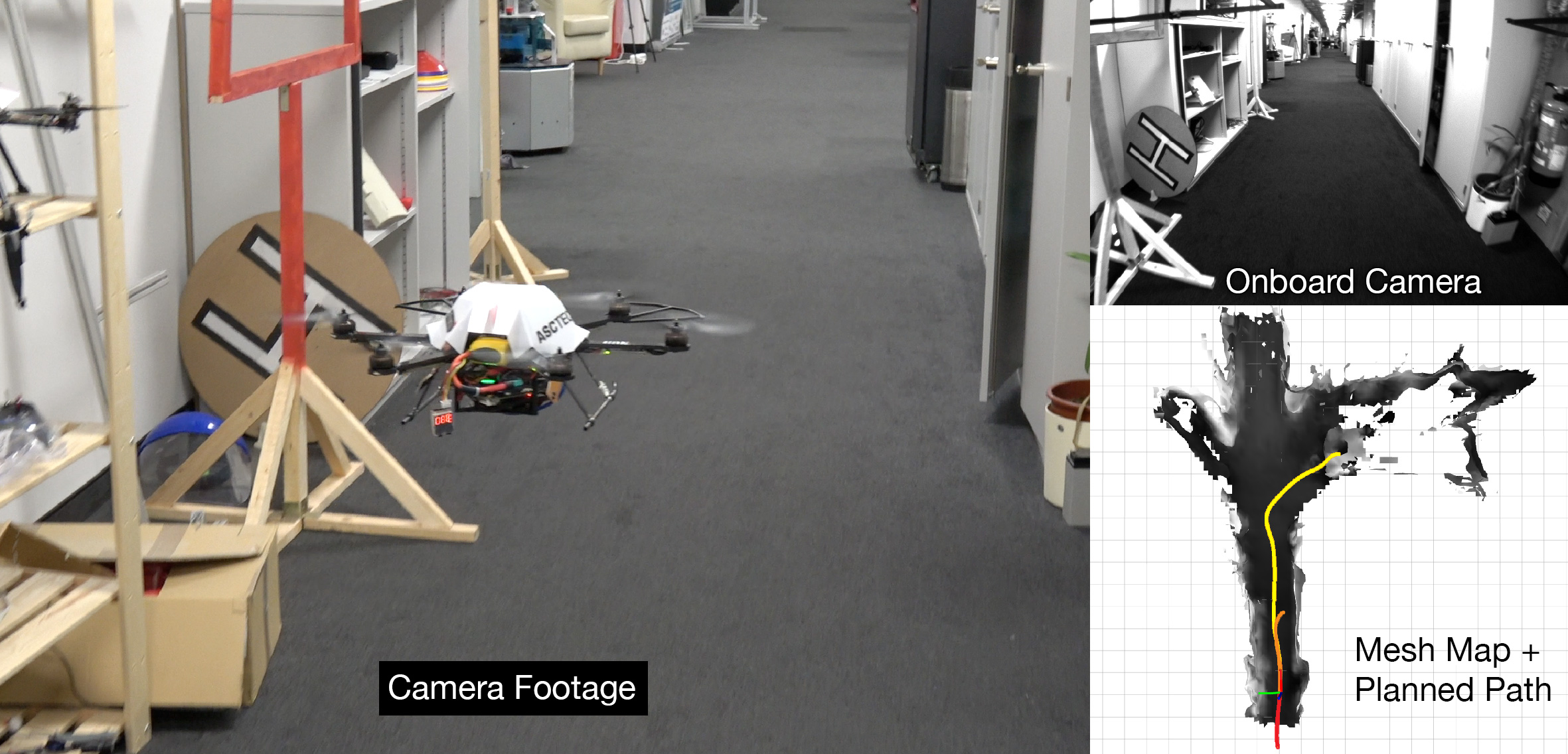}
  \caption{Experimental results from the office navigation experiment, with final map and intermediate paths shown on the lower right.}
  \label{fig:office_exp}
\end{figure}

In the office space environment, the MAV is able to navigate from a starting position in a hallway, around a corner, and to a point above an office table, shown in \reffig{fig:office_exp}.
During the path, it successfully avoids an ajar cabinet door (which blows open during the flight), along with many obstacles on either side of the hallway.
The MAV was only able to reach near the intended goal, as it is unable to successfully determine whether the air-space above the tables is clear or not: the tables are gray and textureless, and the white projector screen behind them is also textureless, leading to a lack of stereo matches and therefore unknown space in the map.
While eventually the robot would have explored enough of the surroundings to clear this space, the pilot intervened when it was near the intended target.
This demonstrates the conservative and safe nature of our planner.

Our second experimental validation took place in a forest environment, where we performed four different experiments.
In the first trial, we were successfully able to avoid a single large tree between the start point and goal. Second, we did two experiments where the MAV was commanded to go a large distance in its current facing direction, where the robot successfully avoided tree branches along its way and navigated largely along a hiking trail for up to 45.0 meters. In the shorter experiment, the MAV was able to reach its goal. In the longer one, it was unable to reach the final goal as the slope of the ground was too high and the tilted-down camera did not allow it to perceive enough open space to safely raise its flying height above the ground.

The final forest experiment tested navigation in very cluttered, obstacle-dense environments.
The MAV was commanded to fly in a very densely-forested area between two trails, containing many small trees, branches, uneven terrain, and other obstacles.
A still image of the video, along with the corresponding robots-eye view and the final executed path are shown in \reffig{fig:forest_exp}.
The MAV was able to complete a path of 34.7 meters, successfully avoiding obstacles along the way, and finishing at the waypoint above the trail on the other side of the wooded region.

\begin{table}[tb]
  \centering
  \begin{tabular}{lr}
    \toprule
    \tableheader{1.5cm}{\textbf{Step}} &  \tableheader{1.5cm}{\textbf{Time} [\textit{ms}]}  \\
    \midrule[1.5pt]
    \multicolumn{2}{l}{\textbf{Mapping}} \\ 
    \midrule
    TSDF Insert & 27.0  \\
    ESDF Update & 14.5 \\ 
    \midrule
    \multicolumn{2}{l}{\textbf{Local Replanning}} \\
    \midrule
    Trajectory Optimization & 19.3  \\
    Intermediate Goal Selection & 5.9 \\
    \bottomrule
  \end{tabular}
  \caption{Timings for a single iteration each part of the method, aggregated from the benchmark in \reffig{fig:long_benchmark}. Note that intermediate goal selection will only run if trajectory optimization fails, not every planning iteration.}
  \label{table:timing}
\end{table}


\section{Conclusions}
This paper presented a complete system for local obstacle avoidance, consisting of an underlying trajectory optimization method, which uses an Euclidean Signed Distance Field (ESDF) built by \textit{voxblox} to get collision costs and gradients, coupled with an exploration-inspired intermediate goal finding strategy to escape local minima in the optimization.
We showed that our combined method outperforms the common strategy of coupling an optimistic global planner with a conservative local planner.
In the case of high obstacle densities, our exploration-based method is able to find solutions to more planning problems.
We also outperform the next-best view exploration method for intermediate goal, as we are able to incorporate information about the global goal and reduce the runtime of the exploration gain evaluation.

Our approach focuses on solving the case of very cluttered environments in previously unknown maps, and maximizing the chances of finding the goal while building the map.
To demonstrate the performance of our method in real-world scenarios, we were able to successfully navigate through an office and through multiple forest environments while performing all processing in real-time on-board an MAV.

\bibliographystyle{ieeetr}

\bibliography{sources}

\begin{thebibliography}{10}

\bibitem{chen2016online}
J.~Chen, T.~Liu, and S.~Shen, ``Online generation of collision-free
  trajectories for quadrotor flight in unknown cluttered environments,'' in
  {\em International Conference on Robotics and Automation (ICRA)}, 2016.

\bibitem{pivtoraiko2013incremental}
M.~Pivtoraiko, D.~Mellinger, and V.~Kumar, ``Incremental micro-uav motion
  replanning for exploring unknown environments,'' in {\em IEEE International
  Conference on Robotics and Automation (ICRA)}, pp.~2452--2458, IEEE, 2013.

\bibitem{oleynikova2016continuous-time}
H.~Oleynikova, M.~Burri, Z.~Taylor, J.~Nieto, R.~Siegwart, and E.~Galceran,
  ``Continuous-time trajectory optimization for online uav replanning,'' in
  {\em IEEE/RSJ International Conference on Intelligent Robots and Systems
  (IROS)}, 2016.

\bibitem{usenko2017real}
V.~Usenko, L.~von Stumberg, A.~Pangercic, and D.~Cremers, ``Real-time
  trajectory replanning for mavs using uniform b-splines and 3d circular
  buffer,'' in {\em IEEE/RSJ International Conference on Intelligent Robots and
  Systems (IROS)}, 2017.

\bibitem{dong2016motion}
J.~Dong, M.~Mukadam, F.~Dellaert, and B.~Boots, ``Motion planning as
  probabilistic inference using gaussian processes and factor graphs,'' in {\em
  Proceedings of Robotics: Science and Systems}, June 2016.

\bibitem{karaman2011sampling}
S.~Karaman and E.~Frazzoli, ``Sampling-based algorithms for optimal motion
  planning,'' {\em The International Journal of Robotics Research}, vol.~30,
  no.~7, pp.~846--894, 2011.

\bibitem{bircher2016receding}
A.~Bircher, M.~Kamel, K.~Alexis, H.~Oleynikova, and R.~Siegwart, ``Receding
  horizon ``next-best-view" planner for 3d exploration,'' in {\em IEEE
  International Conference on Robotics and Automation (ICRA)}, IEEE, 2016.

\bibitem{oleynikova2017voxblox}
H.~Oleynikova, Z.~Taylor, M.~Fehr, R.~Siegwart, and J.~Nieto, ``Voxblox:
  Incremental 3d euclidean signed distance fields for on-board mav planning,''
  in {\em IEEE/RSJ International Conference on Intelligent Robots and Systems
  (IROS)}, 2017.

\bibitem{oleynikova2015reactive}
H.~Oleynikova, D.~Honegger, and M.~Pollefeys, ``Reactive avoidance using
  embedded stereo vision for mav flight,'' in {\em IEEE International
  Conference on Robotics and Automation (ICRA)}, IEEE, 2015.

\bibitem{florence2016integrated}
P.~Florence, J.~Carter, and R.~Tedrake, ``Integrated perception and control at
  high speed: Evaluating collision avoidance maneuvers without maps,'' in {\em
  Workshop on the Algorithmic Foundations of Robotics}, 2016.

\bibitem{lopez2017aggressive}
B.~T. Lopez and J.~P. How, ``Aggressive 3-d collision avoidance for high-speed
  navigation,'' in {\em IEEE International Conference on Robotics and
  Automation (ICRA)}, IEEE, 2017.

\bibitem{richter2013polynomial}
C.~Richter, A.~Bry, and N.~Roy, ``Polynomial trajectory planning for aggressive
  quadrotor flight in dense indoor environments,'' in {\em Proceedings of the
  International Symposium on Robotics Research (ISRR)}, 2013.

\bibitem{burri2015real-time}
M.~Burri, H.~Oleynikova, , M.~W. Achtelik, and R.~Siegwart, ``Real-time
  visual-inertial mapping, re-localization and planning onboard mavs in unknown
  environments,'' in {\em IEEE/RSJ International Conference on Intelligent
  Robots and Systems (IROS)}, Sept 2015.

\bibitem{ratliff2009chomp}
N.~Ratliff, M.~Zucker, J.~A. Bagnell, and S.~Srinivasa, ``Chomp: Gradient
  optimization techniques for efficient motion planning,'' in {\em IEEE
  International Conference on Robotics and Automation (ICRA)}, IEEE, 2009.

\bibitem{heng2014autonomous}
L.~Heng, D.~Honegger, G.~H. Lee, L.~Meier, P.~Tanskanen, F.~Fraundorfer, and
  M.~Pollefeys, ``Autonomous visual mapping and exploration with a micro aerial
  vehicle,'' {\em Journal of Field Robotics}, vol.~31, no.~4, pp.~654--675,
  2014.

\bibitem{shen2012autonomous}
S.~Shen, N.~Michael, and V.~Kumar, ``Autonomous indoor 3d exploration with a
  micro-aerial vehicle,'' in {\em IEEE International Conference on Robotics and
  Automation (ICRA)}, pp.~9--15, IEEE, 2012.

\bibitem{charrow2015information}
B.~Charrow, G.~Kahn, S.~Patil, S.~Liu, K.~Goldberg, P.~Abbeel, N.~Michael, and
  V.~Kumar, ``Information-theoretic planning with trajectory optimization for
  dense 3d mapping.,'' in {\em Robotics: Science and Systems}, 2015.

\bibitem{papachristos2017uncertainty-aware}
C.~Papachristos, S.~Khattak, and K.~Alexis, ``Uncertainty-aware receding
  horizon exploration and mapping using aerial robots,'' in {\em IEEE
  International Conference on Robotics and Automation (ICRA)}, 2017.

\bibitem{davis2016c-opt}
B.~Davis, I.~Karamouzas, and S.~J. Guy, ``C-opt: Coverage-aware trajectory
  optimization under uncertainty,'' {\em IEEE Robotics and Automation Letters},
  2016.

\bibitem{mellinger2011minimum}
D.~Mellinger and V.~Kumar, ``Minimum snap trajectory generation and control for
  quadrotors,'' in {\em IEEE International Conference on Robotics and
  Automation (ICRA)}, pp.~2520--2525, IEEE, 2011.

\bibitem{nieuwenhuisen2016layered}
M.~Nieuwenhuisen and S.~Behnke, ``Layered mission and path planning for mav
  navigation with partial environment knowledge,'' in {\em Intelligent
  Autonomous Systems 13}, pp.~307--319, Springer, 2016.

\bibitem{bloesch2015robust}
M.~Bloesch, S.~Omari, M.~Hutter, and R.~Siegwart, ``Robust visual inertial
  odometry using a direct ekf-based approach,'' in {\em IEEE/RSJ International
  Conference on Intelligent Robots and Systems (IROS)}, pp.~298--304, IEEE,
  2015.

\bibitem{kamel2016linear}
M.~Kamel, M.~Burri, and R.~Siegwart, ``Linear vs nonlinear mpc for trajectory
  tracking applied to rotary wing micro aerial vehicles,'' {\em arXiv preprint
  arXiv:1611.09240}, 2016.

\end{thebibliography}

\end{document}